# Deep Learning on Retina Images as Screening Tool for Diagnostic Decision Support

Maria Camila Alvarez Triviño[1], Jérémie Despraz[2], Jesús Alfonso López Sotelo[1] and Carlos Andrés Peña[2]

[1]*Engineering faculty, Universidad Autónoma de Occidente, Cali, Colombia*
[2]*School of Business and Engineering Vaud (HEIG-VD), University of Applied Sciences of Western Switzerland (HES-SO), Yverdon-les-Bains, Switzerland*
{mcalvarez, jalopez}@uao.edu.co, {jeremie.despraz, carlos.pena}@heig-vd.ch

Keywords: Deep-Learning, Convolutional Neural Networks, Diabetic Retinopathy, Image Processing, Medical Imaging.

Abstract: In this project, we developed a deep learning system applied to human retina images for medical diagnostic decision support. The retina images were provided by EyePACS (Eyepacs, LLC). These images were used in the framework of a Kaggle contest (Kaggle INC, 2017), whose purpose to identify diabetic retinopathy signs through an automatic detection system. Using as inspiration one of the solutions proposed in the contest, we implemented a model that successfully detects diabetic retinopathy from retina images. After a carefully designed preprocessing, the images were used as input to a deep convolutional neural network (CNN). The CNN performed a feature extraction process followed by a classification stage, which allowed the system to differentiate between healthy and ill patients using five categories. Our model was able to identify diabetic retinopathy in the patients with an agreement rate of 76.73% with respect to the medical expert's labels for the test data.

## 1 INTRODUCTION

Nowadays, medical images can be used to assess the health of a patient, enabling clinicians to perform diagnostics and start the corresponding treatment for each pathology; such as breast cancer, kidney deficiencies, diabetic retinopathy, among others. Applying computer aided diagnostic (CAD) in the healthcare industry can help to automate the decision making process, enhance the visualization of data, and improve the extraction of complex features from medical images. Because of the above, CAD can greatly beneficiate medical experts in making clinical diagnostics. CAD uses different machine learning techniques such as deep learning, a technique which generally requires large amount of data in order to be able to extract relevant information and automatically highlight the specific characteristics that differentiate each pathology.

Diabetic retinopathy (DR) is a complication of diabetes that has become one of the leading causes of blindness in working age adults (CDC, Centers for Disease Control and Prevention). It is characterized by a deterioration of the retinal blood vessels that can cause them to swell and leak blood into the vitreous or, in advanced stages, it can produce an abnormal growth of new blood vessels. In this project, we developed an automated detection system for DR based on image pre-processing and deep learning.

## 2 RELATED WORK

Deep learning is a discipline that allows the development of algorithms capable of learning by themselves from a large set of data, without the need of being programed explicitly. In 2016, Google researchers developed an automated system using deep learning that was able to detect diabetic retinopathy (DR) and macular edema in colour retinal fundus photographs (Gulshan, Peng, Coram, & Stumpe, 2016). This method, thanks to an automated system for DR detection has different potential benefits such as increased efficiency, reproducibility, and coverage of the screening programs around the word. As a consequence of these novel techniques, it is hoped that the blindness due to DR can be reduced through early detection and treatment. Researchers used the kind of artificial neural networks optimized for image classification, namely deep convolutional neural network, training it with 128.175 labeled retinal images. These images were provided by EyePACS (Eyepacs, LLC) and three eye hospitals in India (Aravind Eye Hospital, Sankara Nethralaya, and Narayana Nethralaya). The images were rated between 3 to 7 times for diabetic retinopathy, diabetic macular edema, and image gradability by a group of 54 US-licensed ophthalmologists or ophthalmology trainees in their last year of residency. The diabetic retinopathy severity was graded in agreement with the International Clinical DR scale. According to this scale, the levels are: none, mild, moderate, severe, and proliferative DR. Figure 1 illustrates an example of retinal images with its corresponding DR grade, and at the bottom the mentioned scale itself.

The optimization algorithm used to train the network weights was distributed stochastic gradient descent. To speed up the learning phase, they implemented batch normalization and pre-initialization of the weights with the same network trained for object classification with the ImageNet dataset. The network's performance was measured as the area under the receiver operating curve (AUC) resulting from plotting sensitivity vs specificity. Predictions were made using an ensemble of 10 networks trained with the same data, and the final prediction was computed by a linear average over the ensemble predictions.

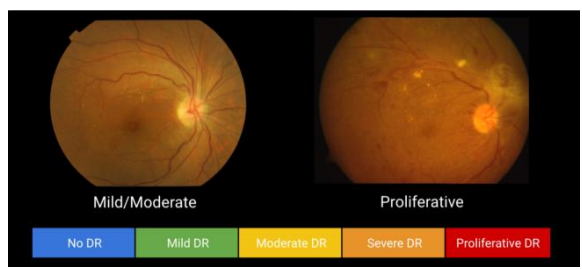

Figure 1: International Clinical diabetic retinopathy scale, image from (Gulshan, Peng, Coram, & Stumpe, 2016) .

The resulting algorithm was validated at the beginning of 2016 using two separated datasets: EyePACS-1 and Messidor-2. Both datasets were evaluated by at least 7 US board-certified ophthalmologists (ABMS), with high intragrader consistency. The algorithm was evaluated through the use of two operating points selected from the development set, one selected for high specificity and the other for high sensitivity. Using the first operating point, high specificity, for EyePACS-1 the algorithm's sensitivity was 90.3% and the specificity was 98.1%. In Messidor-2, the values were 87.0% and 98.5% for sensitivity and specificity respectively. Using the second operating point, high sensitivity on the development set, for EyePACS-1 the values were 97.5% for sensitivity and 93.4% for specificity. While, in Messidor-2 the sensitivity was 96.1% and the specificity 93.9%. In spite of these promising results, it was concluded that more investigation is needed to evaluate the viability of applying the developed algorithm in a clinical environment or to conclude that this technique would represent a significant improvement compared with the care and outcomes of the current ophthalmologic assessments.

# 3 METHODS

## 3.1 Data

This project used two datasets. The first dataset was downloaded from the crowd-sourcing platform Kaggle (Kaggle INC, 2017), in which a Diabetic Retinopathy Detection contest was held. These images were provided by EyePACS (Eyepacs, LLC). The dataset consisted of 88.000 high resolution retina images (training and test), taken under various conditions. A clinician evaluated the images rating them with a DR level of 0 to 4, corresponding to the International Clinical DR scale showed in Figure 1.

The second dataset was provided by the Lausanne University Hospital (CHUV, Centre hospitalier universitaire vaudois) and consisted of a small set of retina images taken in a sample of individuals from the Swiss population in the context of an epidemiological study.

## 3.2 Preprocessing

Because of the different conditions in which the images were obtained, a preprocessing stage was required in order to obtain a uniform set of inputs. Images were cropped in such a way that the retina was positioned at the center and occupy as much space as possible, thus reducing the amount of black pixels corresponding to the background. This was achieved by removing pixel rows and lines of the images that did not sum up above a given threshold, starting from the center and moving towards the margins.

Afterwards, using methods described in (Graham, 2015), we subtracted the local average color from each pixel. The average local color of each pixel was obtained by passing the image through a Gaussian filter (Jain & Kasturi, 1995). This removed most of the background color and highlighted finer details of the image such as veins and stains (see Figure 3). Afterwards, we computed the retina diameter by comparing the pixel values in a line crossing the middle of the image. This allowed us to compute a circular mask that we used to segment the original image into background and retina regions. Note that, in order to remove boundaries effects on the outer regions of the retina, the mask area was reduced by 10% of its original size. The entire preprocessing is summarized schematically in Figure 2.

Finally, the images were re-scaled to a size of 512x512 while maintaining the same aspect ratio and the background pixels were mapped to 50% grey level.

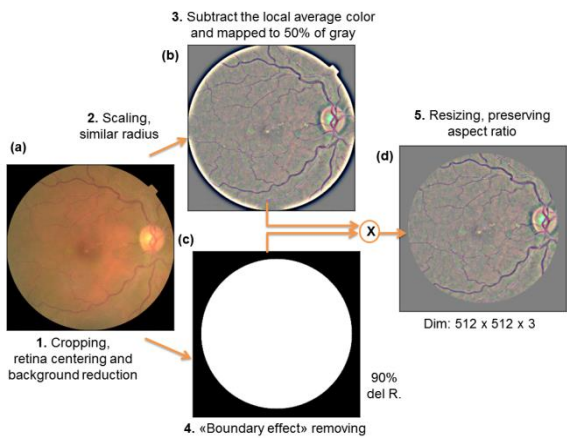

Figure 2: Summary of image preprocessing steps.

### 3.3 Implementation

As starting point, we used the architecture proposed by the DeepSense team (Deepsense.Io, 2015) as solution for the Kaggle's contest. This architecture was then modified to obtain our final best performing model whose architecture is presented in Table 1.

Table 1: Detailed CNN architecture

| LAYER TYPE | OUTPUT SHAPE |
|---|---|
| Convolution 2D | 16, 512, 512 |
| Batch Normalization | |
| Convolution 2D | 16, 512, 512 |
| Batch Normalization | |
| Max pooling | 16, 255, 255 |
| Convolution 2D | 32, 255, 255 |
| Batch Normalization | |
| Convolution 2D | 32, 255, 255 |
| Batch Normalization | |
| Max pooling | 32, 127, 127 |
| Convolution 2D | 64, 127, 127 |
| Batch Normalization | |
| Convolution 2D | 64, 127, 127 |
| Batch Normalization | |
| Max pooling | 64, 63, 63 |
| Convolution 2D | 96, 63, 63 |
| Batch Normalization | |
| Max pooling | 96, 31, 31 |
| Convolution 2D | 96, 31, 31 |
| Batch Normalization | |
| Max pooling | 96, 15, 15 |
| Convolution 2D | 128, 15, 15 |
| Batch Normalization | |
| Max pooling | 128, 7, 7 |
| Dropout | |
| Flatten | 6272 |
| Dense + Regularizer | 96 |
| Dropout | |
| Batch Normalization | |
| Dense | 5 |
| softmax | |

The methodology used to find the model which best fits the problem was through experimentation, modifying the model's hyperparameters according to the issues that occurred after observation (e.g. overfitting, strong class imbalance) in order to maximize K for the test data. These modifications consisted in adding L2 regularizers, batch normalization layers and additional dropout layers to reduce the overfitting.

The deep-learning software library used in this project was Keras (Chollet, 2017) with Theano (LISA Lab, 2017) as backend, and the training of the network itself was preformed on a GeForce GTX 980 Ti. The metric used to evaluate the network's performance was the quadratic weighted kappa (K), (Cohen, 1968). This metric has the advantage of penalizing correct but random guesses by taking into account the labels distribution.

### 3.4 Balancing Data Distribution

The main issue with the available data was the strong class imbalance between healthy and ill subjects, the former being overrepresented in the samples. In order to solve this issue, we used data augmentation techniques.

Since images were analyzed by batches, we constructed these batches in such a way that their inner class distribution was uniform. This was achieved by implementing a custom image selection process to fill these batches. The image selection was based on randomly selecting images from the entire dataset with a probability inversely proportional to their original distribution.

In addition, we applied random rotations of angles between 0-360 deg and performed horizontal and/or vertical flips with a probability of 50% to artificially augment the data, thus avoiding the issue of having several time the exact same image in the batch.

# 4 RESULTS AND DISCUSSION

## 4.1 EyePACS' Dataset

The preprocessing allowed the retina to be well centred on the image and occupy as much space as possible. On the other hand, local mean substraction highlighted the specific details that can be signs of the pathology such as haemorrhage as shown in the proliferative DR case of Figure 3.

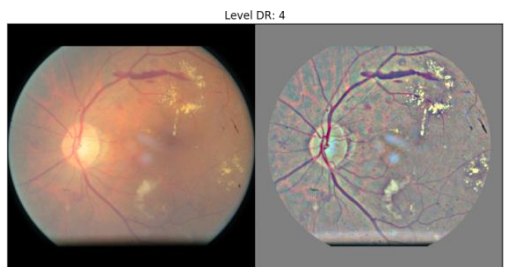

Figure 3: Comparison between an image without and with preprocessing.

Our first experiments demonstrated the importance of balancing the class distribution. As an example, if the data is introduced into the network in batches without considering class distribution, this generates results appearing to be good in terms of training and validation scores in the accuracy vs. epochs graph. However, the kappa value quickly decreases to zero, as the networks starts to strongly overfit the majority class of the dataset. As aforementioned, the above is caused by a strong class imbalance: the zero class (healthy patients) is represented by 74% of the data, while classes three and four, corresponding to high illness levels, are merely represented by 4,52% of the data. This causes the network to obtain a high accuracy (74%) through the trivial solution of labelling all images as belonging to the class which is most frequent in the data.

This also demonstrates the usefulness of the weighted kappa metric that allows for a more accurate assessment of the models performance, since it takes into account the probability of the agreement occurring by chance, producing a score of zero for the above described example.

This motivated and justified the use of dynamic data re-sampling as detailed in section 3.4, which allowed the network to always receive a batch with uniformly distributed classes. However, this re-sampling method made the images belonging to the less represented classes be repeated multiple times. Our experiments demonstrated that this can lead the network to unwanted overfitting, given that it can memorize the training samples rather than actually learn from them. It appeared clearly in the huge difference between the kappa value for the training data (0,96), and the validation dataset (0,25). To solve this issue, we implemented different techniques: augmentation of the amount of dropout layers and dropout probability, addition of regularizer layers, and artificial data augmentation process as mentioned in section 3.4.

Multiple experiments and network configurations where explored in order to find the best model for our problem. The solution which obtained the highest K score for the test data, illustrated in Table 1, had the following architecture: nine convolutional layers, six max pooling layers, batch normalization after each convolutional layer, two fully connected (FC) layers, L2 regularization, and dropout layers of 25% before each fully connected layer. The activation functions were ReLu for the convolutional layers and the first FC layer, and Softmax for the last layer. The loss function used during training was the categorical_crossentropy and the optimizer was adadelta. While K (Kappa) was used as the indicator to measure the network's performance, it was not used directly as a cost function because, in our experiments, it was noticed that such an implementation did not allow for an optimal convergence of the model.

Due to memory limitations, it was necessary to use a Python generator to load the data batch by batch to be sent to the network. In this generator, we implemented the operations of data augmentation described in section 3.4. The overall amount of parameters was 927.911 and the training took around 36 hours in total.

The Kappa's behavior for our best model in its first epochs can be observed in Figure 4. We observe in particular how both the training and validation curves grow together. In total, 144 epochs were ran but are not shown in the graph because the training process was stopped and restarted from the saved weights in different occasions due to technical issues. With this network, the value of K for the test data was 0,70309.

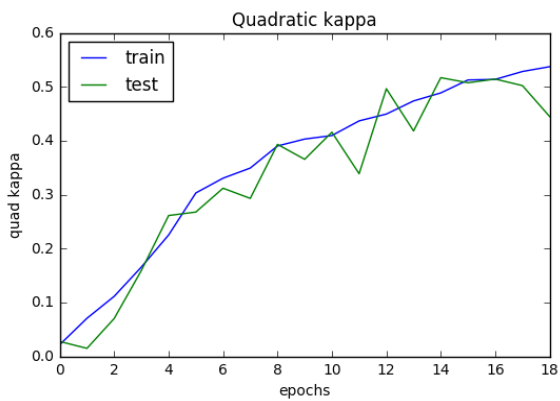

Figure 4: Kappa's behavior with respect to the epochs for our best model.

To increase the Kappa value further, a fine tuning of the trained network's results was performed. During fine tuning, the dynamic re-sampling method was omitted in order to train the network with the original class distribution. With this modification, the K for the test data rose to 0,75265.

To further increase the predictive accuracy of our model, we took advantage of the correlation of images coming from the same patient. Inspired in DeepSesnse's solution, we implemented a post-processing which took into account the probabilities predicted for each eye (left and right) before outputting a final correlated prediction. With this additional step, the final K rose to 0,76736, representing a high level of agreement between the clinical and the automatic system predictions.

Table 2 summarizes the K values obtained on the test set in all of the stages of the final model.

Table 2. Test K summary for each final model stage

| MODEL STAGE | K EVALUATED IN TEST DATA |
|---|---|
| 144 training epochs | 0,70309 |
| Fine tuning | 0,75265 |
| Post-processing analysing both eyes | 0,76736 |

### 4.1 CHUV's Dataset

Finally, we used the CHUV's dataset to qualitatively explore the network's performance in situations close to clinical conditions. 192 images corresponding to 79 patients were introduced into the neural network. The network classified 8 of these images with a DR level different from zero. Visually analyzing this unlabeled images, we found that these individuals indeed had signs similar to those present in DR patient's images. However, since we did not have experts to label precisely the presence or absence of DR, we could not guarantee that these images came from patients with DR, as the signs may also come from other retinopathy types such as hypertensive or pigmentary retinopathy.

In the Figure 5 we can observe some examples of patients' retina images that the network identified with a DR level different from zero. The image in Figure (a) was classified as DR level 2, we observe little yellow points that could be signs of the illness, like hard exudates. In the same way, we see in Figure (b), another retina image classified as DR level 4, i.e. proliferative DR; in this image we can see the presence of what could be a blood stain that could be indicative of a vitreous hemorrhage.

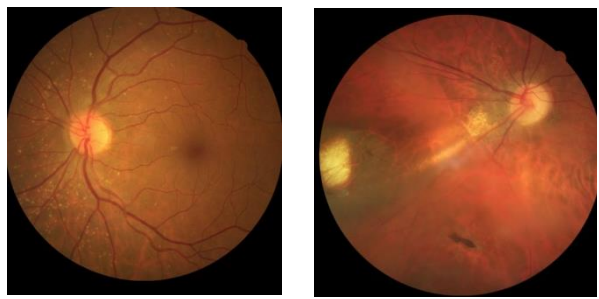

(a) DR class: 2     (b) DR class: 4

Figure 5: Patients' retina images with predicted DR level different from zero

## 5 CONCLUSION

This project presented the development of an automated system for diabetic retinopathy detection in color retina images, through the implementation of deep learning techniques. The adopted deep learning tool was a deep convolutional neural network, trained with a large set of pre-processed images. Quadratic weighted Kappa values for our best model, evaluated with the test data was 76,74%, representing a good strength of agreement between the predicted and the expected grading. The GitHub repository with the code can be reached at: https://github.com/mcamila777/DL-to-retina-images.